\documentclass[
]{ceurart}

\usepackage{listings}
\usepackage{textcomp}
\usepackage{latexsym}
\usepackage{amssymb}
\usepackage{amsmath}
\usepackage{amsthm}
\usepackage{booktabs}
\usepackage{enumitem}
\usepackage{graphicx}
\usepackage{color}
\newcommand{\textproc}[1]{\textsc{#1}}
\usepackage{algorithm}
\usepackage{algorithmic}
\usepackage{subcaption}
\usepackage{booktabs,tabularx,multirow,array,threeparttable}
\newcolumntype{Y}{>{\centering\arraybackslash}X}
\usepackage{listings}
\begin{document}

\copyrightyear{2025}
\copyrightclause{Copyright for this paper by its authors.
  Use permitted under Creative Commons License Attribution 4.0
  International (CC BY 4.0).}

\conference{Identity-Aware AI workshop at 28th European Conference on Artificial Intelligence,
  October 25, 2025, Bologna, Italy}

\title{MetaRAG: Metamorphic Testing for Hallucination Detection in RAG Systems}

\tnotemark[1]
% \tnotetext[1]{You can use this document as the template for preparing your
%   publication. We recommend using the latest version of the ceurart style.}

% --- Authors -------------------------------------------------
\author[1,2]{Channdeth Sok}[%
  orcid=0009-0008-4547-5946,
  email=channdeth.sok@forvia.com
]
\cormark[1]    % corresponding author mark
% \fnmark[1]     % equal contribution mark (if you want it)

\author[1]{David Luz}

\author[1]{Yacine Haddam}

% --- Affiliations -------------------------------------------
\address[1]{Forvia Paris Tech Center, GIT, Immeuble Lumière, 40 avenue des Terroirs de France, 75012 Paris, France}
\address[2]{ENSAE Paris, Institut Polytechnique de Paris}
% --- Footnotes for marks ------------------------------------
\cortext[1]{Corresponding author.}
% \fntext[1]{These authors contributed equally.}


\begin{abstract}
   Large Language Models (LLMs) are increasingly deployed in enterprise applications, yet their reliability remains limited by \textit{hallucinations}, i.e., confident but factually incorrect information. Existing detection approaches, such as SelfCheckGPT and MetaQA, primarily target standalone LLMs and do not address the unique challenges of Retrieval-Augmented Generation (RAG) systems, where responses must be consistent with retrieved evidence. We therefore present \textbf{MetaRAG}, a \textbf{meta}morphic testing framework for hallucination detection in \textbf{R}etrieval-\textbf{A}ugmented \textbf{G}eneration (RAG) systems. MetaRAG operates in a real-time, unsupervised, black-box setting, requiring neither ground-truth references nor access to model internals, making it suitable for proprietary and enterprise deployments. The framework proceeds in four stages: (1) decompose answers into atomic factoids, (2) generate controlled mutations of each factoid using synonym and antonym substitutions, (3) verify each variant against the retrieved context (synonyms are expected to be entailed and antonyms contradicted), and (4) aggregate penalties for inconsistencies into a response-level hallucination score. MetaRAG further localizes unsupported claims at the span level, enabling transparent visualization of potentially hallucinated segments and supporting configurable safeguards in sensitive use cases.
 Experiments on a proprietary enterprise dataset demonstrate the effectiveness of \textbf{MetaRAG} for detecting hallucinations and enabling trustworthy deployment of RAG-based conversational agents. We also outline a topic-based deployment design that translates MetaRAG’s span-level scores into identity-aware safeguards; this design is discussed but not evaluated in our experiments. 
\end{abstract}

\begin{keywords}
  Large Language Models \sep
  Retrieval-Augmented Generation \sep
  Hallucination Detection \sep
  Metamorphic Testing \sep
  Trustworthy AI
\end{keywords}

\maketitle

\section{Introduction}
Large Language Models (LLMs) such as GPT-4 and Llama-3 are transforming enterprise applications in healthcare, law, and customer service~\cite{openai2024gpt4technicalreport, meta2024llama3,bommasani2021opportunities}. They power chatbots and virtual assistants that interact in natural language, offering unprecedented convenience and efficiency~\cite{singhal2023medpalm}. However, as these systems move into production, a persistent challenge emerges: \textit{hallucinations}, i.e., responses that are fluent and convincing but factually incorrect or unsupported by evidence~\cite{maynez-etal-2020-faithfulness,10.1145/3571730survey}.

In domains such as healthcare, law, and finance, hallucinations are not merely a nuisance but a critical barrier to reliable adoption, raising concerns about user trust, regulatory compliance, and business risk~\cite{lyu2024trustworthy}. Moreover, hallucinations are not uniformly risky: the same unsupported claim can differentially affect specific populations. In healthcare (e.g., pregnancy/trimester-specific contraindications), migration and asylum (e.g., protections for LGBTQ+ refugees), or labor rights (e.g., eligibility by status), ungrounded spans can cause disproportionate harm. Rather than treating users as homogeneous, hallucination detection methods should make such spans \textit{reviewable} at the factoid level so downstream systems can apply identity-aware policies (e.g., stricter thresholds, forced citations, or escalation to a human) when the topic indicates elevated risk. This perspective connects hallucination detection to identity-aware deployment, where span-level evidence enables topic-conditioned safeguards that reduce disproportionate risk.

Ji et~al.~\cite{10.1145/3571730survey} categorize hallucinations into two types:
\begin{itemize}
    \item \textbf{Intrinsic hallucination}: fabricated or contradictory information relative to the model’s internal knowledge.
    \item \textbf{Extrinsic hallucination}: generated information that conflicts with, misrepresents, or disregards externally provided context or retrieved documents.
\end{itemize}

\begin{figure}[h]
    \centering
    \includegraphics[width=0.6\linewidth]{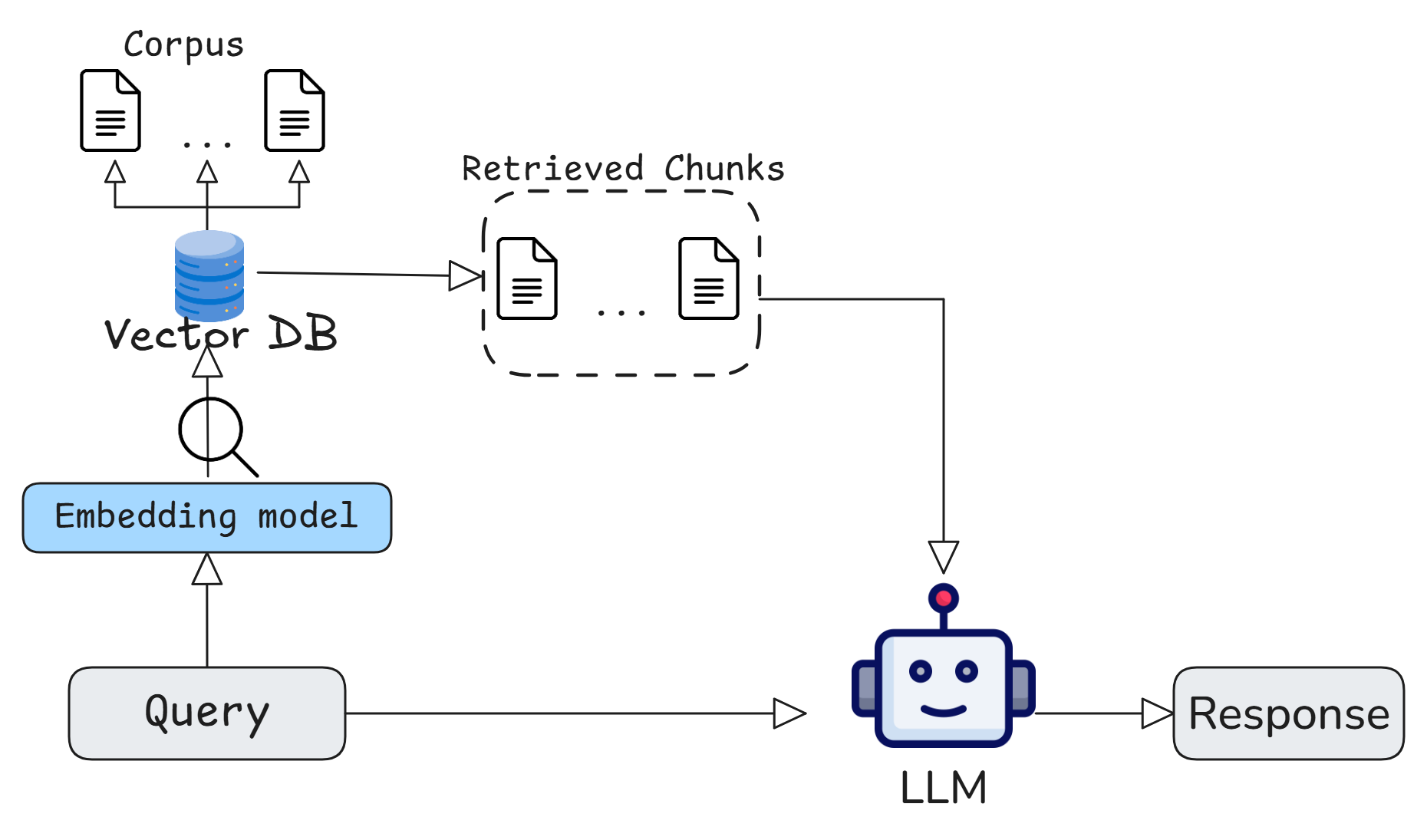}
    \caption{Standard Retrieval-Augmented Generation (RAG) workflow. A user query is encoded into a vector representation using an embedding model and queried against a vector database constructed from a document corpus. The most relevant document chunks are retrieved and appended to the original query, which is then provided as input to a large language model (LLM) to generate the final response.}
    \label{fig:RAGworkflow}
\end{figure}

Retrieval-Augmented Generation (RAG)~\cite{lewis2020rag} aims to mitigate hallucinations by grounding model outputs in retrieved, up-to-date documents, as illustrated in Figure~\ref{fig:RAGworkflow}. By injecting retrieved text from reliable external sources and proprietary documents, into the prompt, RAG improves factuality and domain relevance. While effective against intrinsic hallucinations, RAG remains susceptible to \textit{extrinsic} hallucinations, especially when retrieved evidence is ignored, misinterpreted, or insufficient~\cite{gao2023faithfulrag}.

Detecting hallucinations is particularly challenging in real-world settings, where RAG-based chatbots must respond to queries about unseen, proprietary, or confidential content where gold-standard references are typically unavailable~\cite{liang2022holistic}. Many existing hallucination detection methods rely on gold-standard reference answers~\cite{maynez-etal-2020-faithfulness, kryscinski-etal-2020-evaluating}, annotated datasets~\cite{zhang-etal-2024-benchmarking}, or access to model internals such as hidden states or token log-probabilities~\cite{rashkin-etal-2023-measuring, dziri-etal-2022-faithdial}. However, in enterprise settings, such internals are often inaccessible: many state-of-the-art LLMs (e.g., GPT-4, Claude) are proprietary and only accessible via APIs that expose the final output text but not intermediate computations, limiting the feasibility of these methods in practice~\cite{liang2022holistic}.

% which limits their applicability in enterprise settings where such resources are often unavailable~\cite{liang2022holistic}.

To address these challenges, we introduce \textbf{MetaRAG}: a metamorphic testing framework for detecting hallucinations in RAG-based conversational agents. \textbf{MetaRAG is a zero-resource, black-box setting} that decomposes answers into atomic factoids, applies controlled mutations (e.g., synonym and antonym substitutions), and verifies each mutated factoid against the retrieved context. Synonyms are expected to be \textit{entailed}, while antonyms are expected to be \textit{contradicted}. Hallucinations are flagged when outputs violate these well-defined metamorphic relations (MRs). Unlike prior approaches, MetaRAG does not require ground-truth labels, annotated corpora, or access to model internals, making it suitable for deployment in proprietary settings.

We evaluate MetaRAG on a proprietary corpus, thus unseen during model training. Our results show that MetaRAG reliably detects hallucinations, providing actionable insights for enhancing chatbot reliability and trustworthiness. These results establish MetaRAG as a practical tool for reliable deployment, and its span-level detection opens the door to identity-aware safeguards.

\noindent\textbf{Our contributions include:}
\begin{itemize}
  \item We introduce \textbf{MetaRAG}, a reference-free, black-box setting, metamorphic testing framework for hallucination detection in RAG systems. It decomposes answers into factoids, applies linguistic transformations (synonym and antonym), and verifies them against retrieved context to produce a hallucination score.
  \item We implement a prototype and evaluate MetaRAG on a proprietary dataset, demonstrating its effectiveness in detecting hallucinations that occur when segments of generated responses diverge from the retrieved context.

  % \item We implement a prototype and evaluate MetaRAG on a proprietary dataset, demonstrating its effectiveness in detecting hallucinations by identifying when and where part of generated responses diverge from retrieved context.
  \item We analyze the performance–latency/cost trade‑offs of MetaRAG and provide a consistency analysis to guide future research and practical deployment.
  \item We outline identity-aware safeguards (topic-aware thresholds, forced citation, escalation) that can consume MetaRAG’s scores; these safeguards are a deployment design and are not part of our empirical evaluation.

\end{itemize}

% Moreover, hallucinations are not uniformly risky: the same unsupported claim can differentially affect specific populations. In healthcare (e.g., pregnancy/trimester-specific contraindications), migration and asylum (e.g., protections for LGBTQ+ refugees), or labor rights (e.g., eligibility by status), ungrounded spans can cause disproportionate harm. Rather than treating users as homogeneous, MetaRAG makes such spans \textit{reviewable} at the factoid level so downstream systems can apply identity-aware policies (e.g., stricter thresholds, forced citations, or escalation to a human) when the topic indicates elevated risk.

\section{Related Works}
\subsection{Definitions of Hallucination}
The term \textit{hallucination} has been used with varying scope across natural language generation tasks. 
Some studies emphasize \textbf{\textit{factuality}}, describing hallucinations as outputs that contradict established facts, i.e., inconsistencies with world knowledge or external ground truth~\cite{wang-etal-2024-factuality,Huang_2025}. 
Others highlight \textbf{\textit{faithfulness}}, where hallucinations occur when generated responses deviate from the user instruction or a reference text, often producing plausible but ungrounded statements particularly in source-conditioned tasks such as summarization or question answering~\cite{rawte2023surveyhallucinationlargefoundation}.
Beyond these two dimensions, researchers also note cases of incoherent or nonsensical text that cannot be clearly attributed to factuality or faithfulness criteria ~\cite{10.1145/3571730survey, maynez-etal-2020-faithfulness}.

Alternative terms have also been introduced. 
\textbf{\textit{Confabulation}} draws on psychology to describe fluent but fabricated content arising from model priors~\cite{sui-etal-2024-confabulation}, while \textbf{\textit{fabrication}} is preferred by some to avoid anthropomorphic connotations~\cite{azamfirei2023hallucinations, maleki2024aihallucinationsmisnomerworth}. 
More recently, Chakraborty et al.~\cite{chakraborty2025} propose a flexible definition tailored to deployment settings, defining a hallucination as \textit{a generated output that conflicts with constraints or deviates from desired behavior in actual deployment, while remaining syntactically plausible under the circumstance.}

\subsection{Hallucination Detection in LLMs}

Building on these definitions, hallucinations have been recognized as a major challenge in text generation. 
Early work in machine translation and abstractive summarization described them as outputs that are not grounded in the input source~\cite{maynez-etal-2020-faithfulness,kryscinski-etal-2020-evaluating,koehn-knowles-2017-six}, motivating the development of evaluation metrics and detection methods for faithfulness and factual consistency across natural language generation tasks.

More recent \textit{reference-free} (unsupervised or
zero-reference) methods aim to detect hallucinations without gold-standard labels by analyzing the model's own outputs. A prominent method is \textbf{SelfCheckGPT}~\cite{manakul2023selfcheckgptzeroresourceblackboxhallucination}, a zero-resource, black-box approach that queries the LLM multiple times with the same prompt and measures semantic consistency across responses. The intuition is that hallucinated content often leads to instability under stochastic re-generation; true facts remain stable, while fabricated ones diverge. Manakul et al. show that SelfCheckGPT achieves strong performance in sentence-level hallucination detection compared to gray-box methods, and emphasize that it requires no external database or access to model internals~\cite{manakul2023selfcheckgptzeroresourceblackboxhallucination}. However, SelfCheckGPT may struggle when deterministic decoding or high model confidence leads to repeating the same incorrect output.

\subsection{Metamorphic Testing}
Metamorphic Testing (MT)~\cite{chen2020metamorphictestingnewapproach} was originally proposed in software engineering to address the \textit{oracle problem} in which the correct output is unknown. MT relies on \textit{metamorphic relations} (MRs): transformations of the input with predictable effects on outputs, enabling error detection without access to ground truth~\cite{7422146}. In machine learning, MT has been applied to validate models in computer vision ~\cite{Dwarakanath_2018} (e.g., rotating an image should not change its predicted class) and NLP~\cite{ribeiro-etal-2020-beyond} 

In hallucination detection for LLMs, \textbf{MetaQA}~\cite{yang2025hallucinationdetectionlargelanguage} leverages MRs by generating paraphrased or antonym-based question variants and verifying whether answers satisfy expected semantic or logical constraints. Relying purely on prompt mutations and consistency checks, MetaQA achieves higher precision and recall than SelfCheckGPT on open-domain QA.

Researchers have also adapted MT for more complex conversational and reasoning settings. \textbf{MORTAR}~\cite{guo2025mortarmultiturnmetamorphictesting} applies dialogue-level perturbations and knowledge-graph-based inference to multi-turn systems, detecting up to four times more unique bugs than single-turn MT. \textbf{Drowzee}~\cite{li2024drowzeemetamorphictestingfactconflicting} uses logic programming to construct temporal and logical rules from Wikipedia, generating fact-conflicting test cases and revealing rates of 24.7\% to 59.8\% across six LLMs in nine domains~\cite{yang2025hallucinationdetectionlargelanguage}.

These works highlight the promise of MT for hallucination detection, but they primarily target open-book QA or multi-turn dialogue, often over short, single-sentence outputs. Prior studies have not addressed hallucination detection in \textit{retrieval-augmented generation} (RAG) scenarios over proprietary corpora, a setting in which ground-truth references are unavailable and model internals are inaccessible. \textbf{MetaRAG} builds on MT by decomposing answers into factoids and designing MRs tailored to factual consistency against retrieved evidence in a zero-resource, black-box setting.

\section{MetaRAG: Methodology}
\subsection{Overview}
\begin{figure}
    \centering
    \includegraphics[width=0.7\linewidth]{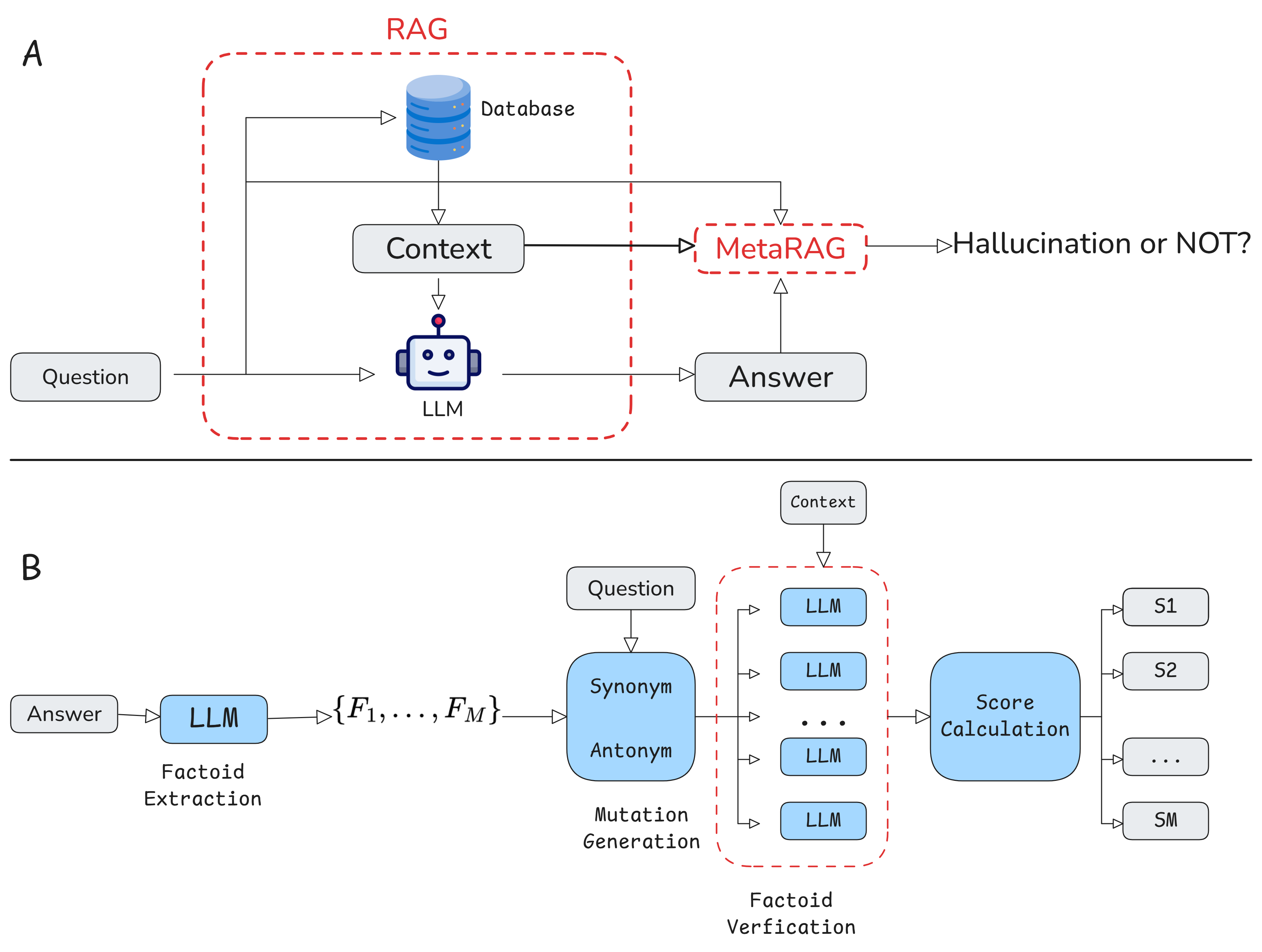}
    \caption{
    Overview of the \textbf{MetaRAG} workflow. 
    \textbf{(A)} Integration of MetaRAG into a standard RAG pipeline: given a user question, the RAG retrieves context and generates an answer, which is then passed to MetaRAG for hallucination detection. 
    \textbf{(B)} Internal MetaRAG pipeline: the answer is decomposed into atomic factoids, each factoid is mutated through synonym and antonym substitutions, and verified against the retrieved context using entailment/contradiction checks. Penalties are assigned to inconsistencies, and scores are aggregated into a response-level hallucination score.
    }
    \label{fig:overviewMetaRAG}
\end{figure}

Building on the metamorphic testing (MT) methodology to detect hallucinations in LLMs introduced by MetaQA~\cite{yang2025hallucinationdetectionlargelanguage}, \textbf{MetaRAG} advances this approach to detect hallucinations in retrieval-augmented generation (RAG) settings by introducing a context-based verification stage. A metamorphic testing layer operates on top of the standard RAG pipeline to automatically detect hallucinated responses. Figure \ref{fig:overviewMetaRAG} outlines the workflow.

Given a user query $Q$, the system retrieves the top-$k$ most relevant chunks from a database, forming the context 
$C = \{c_1, c_2, \ldots, c_k\}$. The LLM generates an initial answer $A$ using $(Q, C)$ as input. 
MetaRAG then decomposes $A$ into factoids, applies controlled metamorphic transformations to produce variants (synonym and antonym), verifies each variant against $C$, and aggregates the results into a hallucination score (Algorithm~\ref{alg:metarag}).

\begin{algorithm}[ht]
\caption{MetaRAG Hallucination Detection}
\label{alg:metarag}
\begin{algorithmic}[1]

\STATE \textbf{Input:}  Generated answer $A$, query $Q$, context $C$, number of mutations $N$, threshold $\tau$
\STATE \textbf{Output:} Hallucination score $H(Q,A,C)$, factoid scores $\{S_i\}$

\STATE \textbf{Factoid Extraction:}
% \STATE $F \leftarrow$ \textproc{FactoidsDecomposition}$(A, \mathrm{LLM}_{\text{extract}})$
\STATE $\mathcal{F} \leftarrow$ \textproc{FactoidsDecomposition}$(A)$
\FOR{each factoid $F_i$ in $\mathcal{F}$}
    \STATE \textbf{Mutation Generation:}
    \STATE $\mathit{Synonyms} \leftarrow$\textproc{GenerateSynonymMutations}$(F_i, Q, N)$
    \STATE $\mathit{Antonyms} \leftarrow$\textproc{GenerateAntonymMutations}$(F_i, Q, N)$
    
    \STATE \textbf{Verification:}
    \FOR{each $F^{syn}$ in $\mathit{Synonyms}$}
        \STATE $\mathit{result} \leftarrow$ \textproc{VerifyWithLLM}$(F^{syn}, C)$
        \STATE $\mathit{SynResults}.\text{append}(\mathit{result})$
    \ENDFOR
    \FOR{each $F^{ant}$ in $\mathit{Antonyms}$}
        \STATE $\mathit{result} \leftarrow$ \textproc{VerifyWithLLM}$(F^{ant}, C)$
        \STATE $\mathit{AntResults}.\text{append}(\mathit{result})$
    \ENDFOR
    
    \STATE \textbf{Scoring:}
    \STATE $\mathit{SynScores} \leftarrow [\textproc{MapSynonymScore}(r)~\text{for}~r~\text{in}~\mathit{SynResults}]$
    \STATE $\mathit{AntScores} \leftarrow [\textproc{MapAntonymScore}(r)~\text{for}~r~\text{in}~\mathit{AntResults}]$
    \STATE $S_i \leftarrow \mathrm{Mean}(\mathit{SynScores} \cup \mathit{AntScores})$
\ENDFOR

\STATE \textbf{Aggregation:}
\STATE $H(Q,A,C) \leftarrow \max_i S_i$
\STATE Return $H(Q,A,C)$, $\{S_i\}$

\end{algorithmic}
\end{algorithm}

\subsection{Step~1: Factoid decomposition}\label{sec:factoid-decom}
Given an answer $A$, we first decompose it into a set of  \emph{factoids}, defined as atomic, independently verifiable facts, denoted by $\mathcal{F}=\{F_1,\dots,F_M\}$. Each factoid $F_j$ corresponds to a single factual statement that cannot be further divided without losing meaning, such as a subject-predicate-object triple or a scoped numerical or temporal claim. Representing an answer $A$ at the factoid level enables fine-grained verification in subsequent steps, allowing localized hallucinations to be marked inside longer answers.

We obtain $\mathcal{F}$ using an LLM-based extractor with a fixed prompt that enforces one proposition per line, prohibits paraphrasing or inference beyond $A$, and co-reference resolution. The full prompt template is provided in the supplementary material.

\subsection{Step~2: Mutation Generation}\label{sec:mutation-generation}

Each factoid (hereafter, \emph{fact}) from Step~1, MetaRAG applies metamorphic mutations to generate perturbed variants of the original claim. This step is grounded in the principle of metamorphic testing, where controlled semantic transformations are used to probe model consistency and expose hallucinations~\cite{yang2025hallucinationdetectionlargelanguage}.

Formally, for each factoid $F_i \in \{F_1, \ldots, F_M\}$, we construct variants using two relations:

\begin{itemize}
    \item \textbf{Synonym Mutation}: This relation substitutes key terms in $F_i$ with appropriate synonyms, yielding paraphrased factoids $F_{i,j}^{\mathrm{syn}}$ that preserve the original semantic meaning. These assess the model’s ability to recognize reworded yet factually equivalent statements.
    \item \textbf{Antonym Mutation}: This relation replaces key terms in $F_i$ with antonyms or negations, producing factoids $F_{i,j}^{\mathrm{ant}}$ that are semantically opposed to the original. These serve as adversarial tests to ensure the model does not support clearly contradictory information.
\end{itemize}

Let $N$ denote the number of mutations generated by \emph{each} relation. The mutation set for $F_i$ is therefore
\[
\mathcal{F}_i=\{\,F_{i,1}^{\mathrm{syn}},\ldots,F_{i,N}^{\mathrm{syn}},\;F_{i,1}^{\mathrm{ant}},\ldots,F_{i,N}^{\mathrm{ant}}\,\}.
\]

By construction, if $F_i$ is correct and supported by the retrieved context $C$, then $F_{i,\cdot}^{\mathrm{syn}}$ should be \emph{entailed} by $C$, whereas $F_{i,\cdot}^{\mathrm{ant}}$ should be \emph{contradicted} by $C$.

Mutations are generated by prompting an LLM with templates that explicitly instruct synonymous or contradictory outputs while preserving atomicity and relevance; the exact prompt templates appear in the supplementary material.

% using LLM prompts that explicitly instruct the model to generate either synonymous or contradictory statements, while preserving atomicity and relevance to the original question. The exact prompt templates are provided in the supplementary material.

\subsection{Step~3: Factoid Verification}\label{sec:factoid-verification}
Each mutated factoid $F_{i,j}^{\mathrm{syn}}$ and $F_{i,j}^{\mathrm{ant}}$ is then verified by LLMs conditioning on the context $C$ (treated as ground truth). The LLM returns one of three decisions:
\textsc{Yes} (entailed by $C$), \textsc{No} (contradicted by $C$), or \textsc{Not sure} (insufficient evidence). 
We then assign a penalty score $p \in \{0, 0.5, 1\}$ based on the decision and the mutation type:

\begin{table}[h]
  \centering
  \caption{Penalty scheme for metamorphic verification (lower is better).}
  \label{tab:penalties}
  \begin{tabular}{lcc}
    \toprule
    & \multicolumn{2}{c}{Penalty $p$}\\
    \cmidrule(lr){2-3}
    Decision & Synonym & Antonym\\
    \midrule
    \textsc{Yes}      & 0.0 & 1.0\\
    \textsc{Not sure} & 0.5 & 0.5\\
    \textsc{No}       & 1.0 & 0.0\\
    \bottomrule
  \end{tabular}
\end{table}
This penalty assignment quantifies semantic (in)consistency at the variant level: correct entailment for synonyms and correct contradiction for antonyms yield zero penalty, while the opposite yields maximal penalty. In Step~4, we aggregate these penalties over all variants of each $F_i$ to compute a fact-level hallucination score.

\subsection{Step 4: Score Calculation}
To quantify hallucination risk, we calculate a hallucination score, $S_i$, for each factoid $F_i$. This yields a granular diagnostic that pinpoints which claims are potentially unreliable. The score for each factoid $i$ is defined as the average penalty across the $2N$ metamorphic transformations (synonym and antonym) of $F_i$:

\begin{equation}
S_i = \frac{1}{2N} \left( \sum_{j=1}^N p_{i,j}^{\mathrm{syn}} + \sum_{j=1}^N p_{i,j}^{\mathrm{ant}} \right),
\end{equation}

where $p_{i,j}^{\mathrm{syn}}$ and $p_{i,j}^{\mathrm{ant}}$ are the penalties assigned in Step~3 to the $j$-th synonym and antonym variants of $F_i$, respectively. By construction, $S_i \in [0,1]$: $S_i=0$ indicates a perfectly consistent, well-grounded factoid, thus no hallucination, while $S_i=1$ indicates a highly probable hallucination.

\paragraph{Response Hallucination score:} 

Instead of a simple average, the hallucination score for the entire response $A$ is defined as the maximum score found among all the individual factoids. This metric ensures that a single, severe hallucination in any part of the response will result in a high overall score, accurately reflecting the unreliability of the entire answer.

\begin{equation}
    H(Q,A,C) \;=\; \max_{1 \le i \le M} S_i,
\end{equation}
where $M$ is the number of decomposed factoids. A response can be flagged as containing hallucination if $H(Q,A,C)$ exceeds a predefined confidence threshold $\tau \in [0,1]$ (e.g. $0.5$).

\subsection{Identity-Aware Safeguards for Deployment}\label{sec:policyhooks}

While MetaRAG is a general-purpose hallucination detector, its
factoid-level scores can be directly integrated into \emph{identity-aware
deployment policies}. Importantly, no protected attributes are inferred
or stored; instead, only the \emph{topic of the query or retrieved context}
(e.g., pregnancy, refugee rights, labor eligibility) is used as a deployment
signal. \emph{Scope.} The safeguards described here represent a deployment design that consumes MetaRAG’s scores; they are not part of the empirical evaluation reported in Section~\ref{sec:experiments}.

Each factoid receives a score $S_i \in [0,1]$, where $S_i=0$ indicates
full consistency with the retrieved context and $S_i=1$ indicates strong
evidence of hallucination. The overall response score $H(Q,A,C)$ thus
represents the risk level of the most unreliable claim: higher values
correspond to higher hallucination risk.

These scores could enable deployment-time safeguards through the following
hooks:

\begin{enumerate}
  \item \textbf{Topic detection.} A lightweight topic classifier or
  rule-based tagger can assign coarse domain labels (e.g., healthcare,
  migration, labor) to the query or retrieved context.

  \item \textbf{Topic-aware thresholds.} A response is flagged if
  $H(Q,A,C) \geq \tau$. Thresholds can be adapted by domain, e.g.,
  $\tau_{\text{general}}=0.5$ for generic queries, and a stricter
  $\tau_{\text{identity}}=0.3$ for sensitive domains.

  \item \textbf{Span highlighting and forced citation.} For flagged
  responses, MetaRAG highlights unsupported spans and enforces
  inline citations to retrieved evidence, to improve transparency
  and calibrate user trust.

  \item \textbf{Escalation.} If hallucinations persist above threshold in
  identity-sensitive domains, the system may abstain, regenerate with a
  stricter prompt, or escalate to human review.

  \item \textbf{Auditing.} Logs of flagged spans, hallucination scores,
  and topic labels can be maintained for post-hoc fairness, compliance,
  and safety audits.
\end{enumerate}

In this way, higher hallucination scores are systematically translated
into stronger protective actions, with more conservative safeguards
applied whenever queries touch on identity-sensitive contexts.
\section{Experiments}\label{sec:experiments}
We conducted experiments to evaluate \textbf{MetaRAG} on its ability to \emph{detect} hallucinations in retrieval-augmented generation (RAG). The evaluation simulates a realistic enterprise deployment setting, in which a chatbot serves responses generated from internal documentation. Our focus is on the detection stage, that is, identifying when an answer contains unsupported (hallucination) or fabricated information. Prevention and mitigation are important but they are outside the scope of this work.

\subsection{Dataset} 
The evaluation dataset is a proprietary collection of \textbf{23 internal enterprise documents}, including policy manuals, procedural guidelines, and analytical reports, none of which were seen during LLM training. Each document was segmented into chunks of a few hundred tokens, and retrieval used cosine similarity over \texttt{text-embedding-3-large}, with the top-$k=3$ chunks appended to each query. 

We then collected a set of user queries and corresponding chatbot answers. Each response was labeled by human annotators as either hallucinated or not, using the retrieved context as the reference. The final evaluation set contains \textbf{67 responses}, of which \textbf{36} are labeled as \textit{not hallucinated} and \textbf{31} as \textit{hallucinated}. 

To preserve confidentiality, we do not release the full annotated dataset. However, the complete annotation guidelines are included in the supplementary material.

\subsection{Evaluation Protocol}

MetaRAG produces \textbf{fine-grained, factoid-level hallucination scores}, whereas the available ground truth labels are at the \textbf{response level}. To align with these existing labels, we evaluate MetaRAG as a binary classifier by thresholding the hallucination score $H(Q,A,C)$ at $\tau = 0.5$. We report standard classification metrics: Precision, Recall, F1 score and accuracy. Latency is also recorded to assess feasibility for real-time deployment.

\subsubsection{Case Studies in Identity-Sensitive Domains}
Beyond quantitative evaluation, we also provide qualitative
illustrations of MetaRAG in identity-sensitive scenarios. To
illustrate how MetaRAG’s span-level scores can enable identity-aware
safeguards without inferring protected attributes, we present two
stylized examples. These are not part of the quantitative evaluation
in Section~\ref{sec:experiments}, but highlight potential deployment
scenarios.

\textbf{Healthcare (pregnancy).} 
A user asks: ``Can pregnant women take ibuprofen for back pain?''
The model answers: ``Yes, ibuprofen is \textbf{safe throughout pregnancy}.'' 
However, the retrieved context specifies that ibuprofen is
contraindicated in the third trimester. MetaRAG flags the span
\textbf{``safe throughout pregnancy''} with a high factoid score
($S_i=0.92$), yielding a response-level score $H=0.92$.
Under the policy hooks described in Section~\ref{sec:policyhooks},
the topic tag \textit{pregnancy} triggers a stricter threshold
($\tau_{\text{identity}}=0.3$, lower than the general case), span
highlighting, a forced citation requirement, and possible escalation
to human review.

\textbf{Migration (refugee rights).} 
A user asks: ``Do LGBTQ+ refugees \textbf{automatically receive protection}
in country~X?'' The model claims that such protections are
\textbf{automatic}, but the retrieved legal text provides no evidence
of this. MetaRAG flags the unsupported claim
\textbf{``automatically receive protection''} with a moderate score ($S_i=0.5$), yielding a response-level score $H=0.5$. Although this score would sit at the decision boundary under a general threshold ($\tau_{\text{general}}=0.5$), the stricter identity-aware threshold ($\tau_{\text{identity}}=0.3$) ensures it is flagged for this case.
Under the policy hooks, the topic tag \textit{asylum/refugee} enforces
citation and may escalate the response to a human reviewer. In a
chatbot deployment, the system would abstain from returning the
unsupported answer and instead notify the user that expert
verification is required.

These qualitative vignettes complement our quantitative evaluation by
showing how MetaRAG’s flagged spans can be turned into concrete
safeguards in identity-sensitive deployments.

\section{Ablation Study}
To understand the contribution of individual design choices, we perform a set of ablation experiments using the private dataset.
\subsection{Ablation Study Design}
We evaluate 26 configurations of MetaRAG, each defined by a combination of:
\begin{itemize}
    \item Number of variants per relation $N\in\{2,5\}$ 
    \item Factoid-decomposition model: \textit{gpt-4.1 or gpt-4.1-mini} from OpenAI
    \item Temperature for mutation generation: $T \in \{0.0, 0.7\}$
    \item Mutation–generation model: \textit{gpt-4.1 or gpt-4.1-mini} 
    \item Verifier model: \textit{gpt-4.1-mini}, \textit{gpt-4.1}, or the \textit{multi} ensemble (gpt-4.1-nano, gpt-4.1-mini, gpt-4.1, Claude Sonnet 4)

\end{itemize}
Since the evaluation task is binary classification, we report \textbf{Precision}, \textbf{Recall}, \textbf{F1 score}, and \textbf{Accuracy}, along with \textbf{latency} (lower is better). 

\begin{figure*}[ht]
    \centering
    \includegraphics[width=0.9\linewidth]{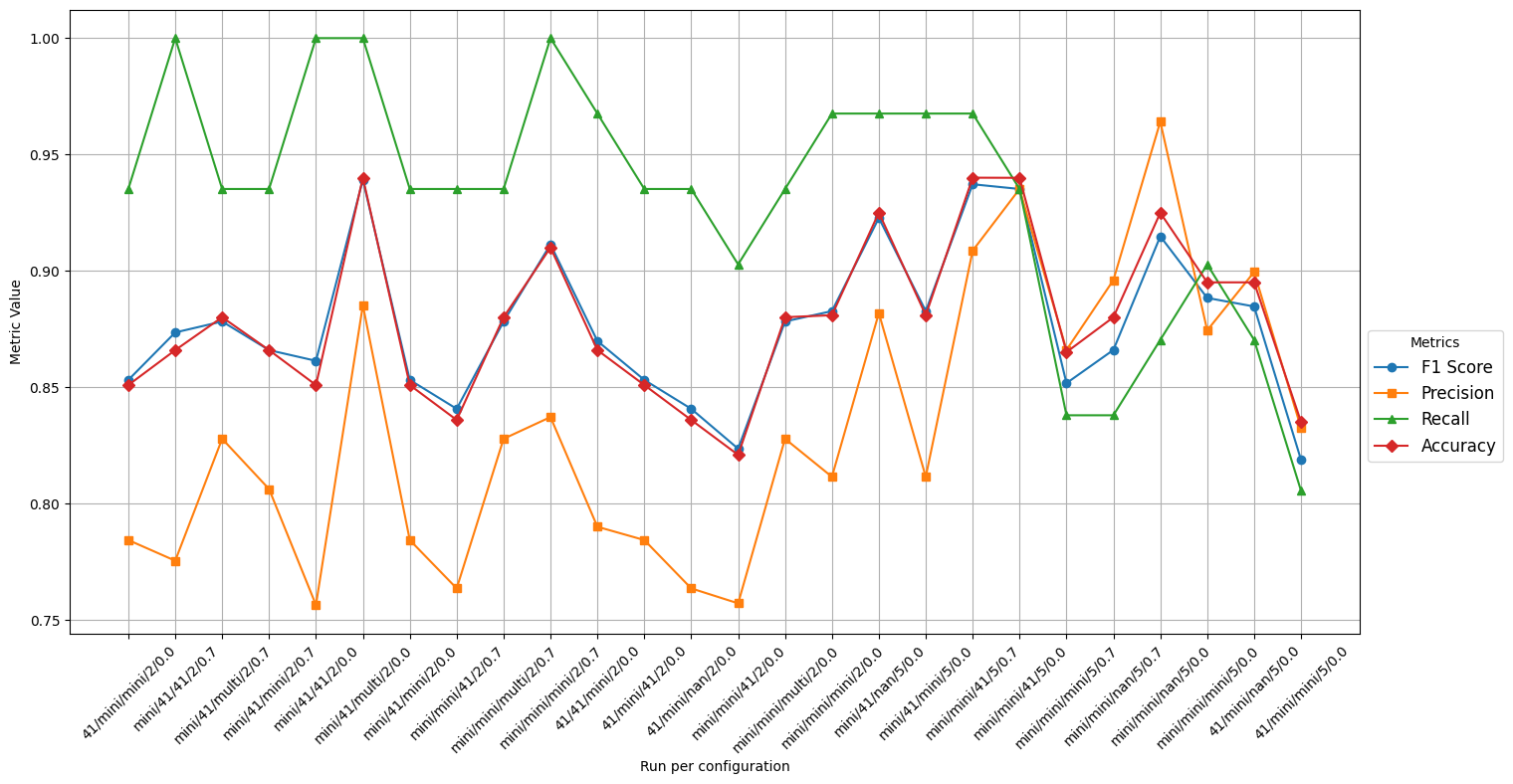}
    \caption{Evaluation metrics for all 26 \textbf{MetaRAG} configurations.}
    \label{fig:plotall}
\end{figure*}

\subsection{Results}

To provide a comprehensive view of performance trade-offs, we report the \textbf{Top-4 configurations separately} for each of three primary metrics: F1 score, Precision, and Recall (Table~\ref{tab:leaderboards}). The configuration notation follows the format: 
\begin{center}
    \texttt{Decomposition Model / Generation Model / Verifier / $N$ / Temperature}. 
\end{center}

For example, \texttt{mini/41/multi/2/0} indicates that the factoid decomposition model is ``mini'', the variant generation model is ``41'', the verifier is ``multi'', there are $N=2$ variants per relation, and the temperature is 0.0.

Several configurations appear in more than one top-4 list, reflecting balanced performance across metrics. For instance, ID 5 (\texttt{mini/41/multi/2/0}) ranks first in both F1 score and Recall, while maintaining competitive Precision.
\begin{table}[ht]
  \caption{Validation leaderboards for 26 \textbf{MetaRAG} configurations, showing the top–4 for each metric.}
  \label{tab:leaderboards}
  \centering
  \footnotesize
  \setlength{\tabcolsep}{3pt} % tighter spacing
  \renewcommand{\arraystretch}{1.05} % compact rows
  \begin{threeparttable}
  \begin{tabular}{
    >{\centering\arraybackslash}p{0.6cm}  % ID
    >{\centering\arraybackslash}p{3.2cm}  % Config
    >{\centering\arraybackslash}p{0.9cm}  % F1
    >{\centering\arraybackslash}p{0.9cm}  % Prec.
    >{\centering\arraybackslash}p{0.9cm}  % Rec.
    >{\centering\arraybackslash}p{0.9cm}  % Acc.
  }
    \toprule
    \textbf{ID} & \textbf{Config.} & \textbf{F1} & \textbf{Prec.} & \textbf{Rec.} & \textbf{Acc.} \\
    \midrule
    \multicolumn{6}{l}{\textbf{Top-4 by F1}} \\
    \cmidrule(lr){1-6}
      5  & mini/41/multi/2/0    & 0.939 & 1.000 & 0.885 & 0.940 \\
      18 & mini/mini/41/5/0.7   & 0.937 & 0.909 & 0.968 & 0.940 \\
      19 & mini/mini/41/5/0     & 0.935 & 0.935 & 0.935 & 0.940 \\
      16 & mini/41/multi/5/0    & 0.923 & 0.882 & 0.968 & 0.925 \\
    \addlinespace[0.4ex]
    \multicolumn{6}{l}{\textbf{Top-4 by Precision}} \\
    \cmidrule(lr){1-6}
      22 & mini/mini/multi/5/0  & 0.915 & 0.964 & 0.870 & 0.925 \\
      19 & mini/mini/41/5/0     & 0.935 & 0.935 & 0.935 & 0.940 \\
      18 & mini/mini/41/5/0.7   & 0.937 & 0.909 & 0.968 & 0.940 \\
      24 & 41/mini/multi/5/0    & 0.885 & 0.900 & 0.870 & 0.895 \\
    \addlinespace[0.4ex]
    \multicolumn{6}{l}{\textbf{Top-4 by Recall}} \\
    \cmidrule(lr){1-6}
      1  & mini/41/41/2/0.7     & 0.874 & 0.776 & 1.000 & 0.866 \\
      5  & mini/41/multi/2/0    & 0.939 & 0.885 & 1.000 & 0.940 \\
      9  & mini/mini/mini/2/0.7 & 0.911 & 0.837 & 1.000 & 0.910 \\
      4  & mini/41/41/2/0       & 0.861 & 0.757 & 1.000 & 0.851 \\
    \bottomrule
  \end{tabular}
  \vspace{0.25ex}
  \begin{tablenotes}[flushleft]\footnotesize
    \item \textbf{Config legend}: \texttt{Decomp/GenModel/Verifier/$N$/Temp}.
  \end{tablenotes}
  \end{threeparttable}
\end{table}

The most promising configurations are further examined in Section~\ref{sec:consistency} to verify stability under multiple seeds.

\subsection{Consistency Check}\label{sec:consistency}
To verify the robustness of our results, each top configuration (selected based on F1 score, Precision, Recall, and token usage) is rerun under identical conditions using five different random seeds. This procedure serves three purposes:  

\begin{itemize}
    \item To ensure that high performance is not attributable to random initialization or favorable seeds.  
    \item To quantify variability across runs with the same configuration by reporting the standard deviation for each metric.  
    \item To assess stability using the coefficient of variation (CV) defined as the ratio of the standard deviation to the mean ($\mathrm{CV} = \sigma / \mu$), where lower values indicate greater consistency.  
\end{itemize}

\begin{table}[ht]
    \centering
    \caption{Run-to-run consistency for top configurations (mean $\pm$ standard deviation over 5 seeds) and coefficient of variation (CV) for F1.}
    \label{tab:cv_table}
    \footnotesize
    \begin{tabular}{lcccc}
        \toprule
        \textbf{ID} & \textbf{F1} & \textbf{Precision} & \textbf{Recall} & \textbf{CV (F1)} \\
        \midrule
        16 & 0.9397 $\pm$ 0.0123 & 0.9198 $\pm$ 0.0243 & 0.9610 $\pm$ 0.0144 & 1.31\% \\
        18 & 0.9356 $\pm$ 0.0089 & 0.9322 $\pm$ 0.0439 & 0.9413 $\pm$ 0.0278 & 0.95\% \\
        19 & 0.9347 $\pm$ 0.0305 & 0.9410 $\pm$ 0.0357 & 0.9286 $\pm$ 0.0272 & 3.26\% \\
        5  & 0.9108 $\pm$ 0.0346 & 0.8463 $\pm$ 0.0503 & 0.9869 $\pm$ 0.0179 & 3.80\% \\
        \bottomrule
    \end{tabular}
\end{table}

Across all metrics, the top configurations demonstrate strong reproducibility, with the majority exhibiting a CV below 2\%. In particular, configurations 18 and 16 achieve both high F1 scores and low variability, indicating that they are not only accurate but also stable across repeated trials.
\begin{figure*}[ht]
  \centering
  \includegraphics[width=0.95\linewidth]{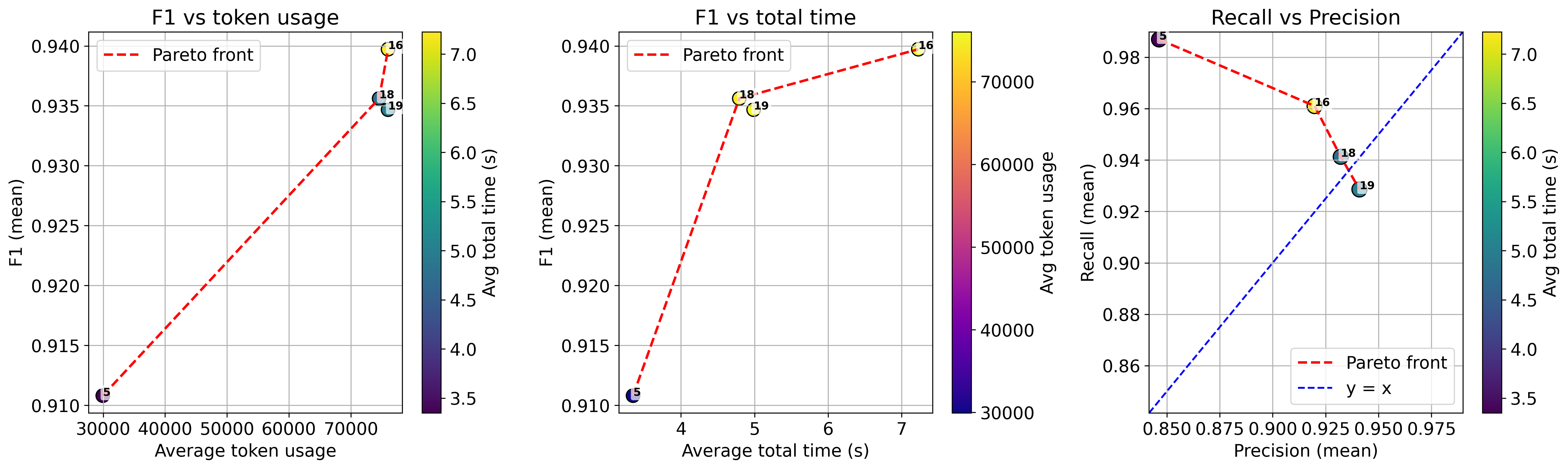}
    \caption{
    Pareto front analysis for hallucination detection performance. 
    Each point represents a MetaRAG configuration; Pareto-optimal points (non-dominated) are highlighted. 
    Subplots show: (Left) F1 vs.\ average token usage, (Center) F1 vs.\ average total execution time, (Right) Precision vs.\ Recall. Pareto-optimal points represent configurations with no strictly better alternative in both accuracy and cost.
    Configuration IDs correspond to Table~\ref{tab:leaderboards}.
  }
  \label{fig:pareto_f1}
\end{figure*}
\subsection{Pareto Front Analysis}\label{sec:pareto}
Following the consistency check (Section~\ref{sec:consistency}), we restrict the Pareto front analysis to the four most stable top-performing configurations selected by F1 score. We analyze the trade-off between hallucination detection performance and efficiency using Pareto frontiers. 
A configuration is \textbf{Pareto-optimal} if no other configuration achieves strictly higher F1 while being no worse in cost metrics; similarly, for precision–recall trade-off.  

Figure~\ref{fig:pareto_f1} presents the Pareto fronts for our primary detection metric (F1 score) with respect to 
(i) average token usage, 
(ii) average total execution time (second), and 
(iii) the precision–recall trade-off.
The Pareto front highlights configurations that offer the best possible balance between accuracy and efficiency, enabling deployment choices aligned with cost or latency constraints.

Several top-ranked configurations (IDs 5, 18, 19, 16) lie on the Pareto front across these views, indicating that they offer competitive accuracy without excessive cost. The corresponding Pareto analyses for precision and recall metrics are provided in the Supplementary Material.

\section{Discussion}
\subsection{Practical Implications}
Integrating hallucination detection into enterprise RAG systems offers several advantages:
\begin{itemize}
    \item \textbf{Risk Mitigation}: Early detection of unsupported answers mitigates the spread of misinformation in both customer-facing and internal applications.
    \item \textbf{Regulatory Compliance}: Many industries, such as healthcare and finance, require verifiable information; automated detection supports regulatory compliance.
    \item \textbf{Operational Efficiency}: Detecting hallucinations simultaneously with content delivery reduces the need for costly downstream human verification.
\end{itemize}

\subsection{Ethical Considerations}

Beyond technical performance, hallucination detection intersects directly with questions of fairness, accountability, and identity harms. 
Hallucinations in chatbot systems pose risks that extend beyond factual inaccuracies: they can reinforce harmful stereotypes, undermine user trust, and misrepresent marginalized communities in identity-sensitive contexts.

\begin{itemize}
    \item \textbf{Reinforced stereotypes:} Language models are known to reproduce and amplify societal biases, as demonstrated by benchmarks such as StereoSet~\cite{nadeem-etal-2021-stereoset} and WinoBias~\cite{zhao-etal-2018-gender}. In identity-sensitive deployments, hallucinated outputs risk reinforcing these biases in subtle but harmful ways.
    \item \textbf{Trust erosion:} Chatbots are only adopted at scale in high-stakes domains if users trust their outputs. Surveys on hallucination consistently highlight that exposure to unsupported or fabricated content undermines user trust in LLM systems~\cite{10.1145/3571730survey,lyu2024trustworthy}.
    \item \textbf{Identity harms:} Misrepresentations in generated responses may distort personal narratives or marginalize underrepresented groups, aligning with broader critiques that technical systems can reproduce social inequities if identity considerations are overlooked~\cite{gebru2021datasheets,selbst2019fairness}.

\end{itemize}

By detecting hallucinations in a black-box, reference-free manner, \textbf{MetaRAG} can support safer deployment of RAG-based systems, particularly in settings where fairness, identity, and user well-being are at stake.

\subsection{Limitations and Future Work}

While MetaRAG demonstrates strong hallucination detection performance, several limitations remain:

\begin{itemize}
    \item \textbf{Dataset Scope:} The study relies on a private, domain-specific dataset. This may limit external validity. \emph{Future work should focus on curating or constructing public benchmarks designed to avoid overlap with LLM pretraining corpora, enabling more robust generalization}.
    \item \textbf{Annotation Granularity:} We lack factoid-level ground truth, which reduces our ability to assess fine-grained reasoning accuracy. \emph{Providing such annotations in future datasets would support deeper consistency evaluations}.
    \item \textbf{Policy Hooks Not Evaluated:} The identity-aware deployment hooks introduced in Section~\ref{sec:policyhooks} are presented only as a design concept. In our implementation, we used a fixed threshold of $\tau=0.5$ across all queries. \emph{Future research should implement and measure the effectiveness of topic-aware thresholds, forced citation, and escalation strategies in real-world chatbot deployments}.
    
    \item \textbf{Topic as Proxy (Design Limitation):} In Section~\ref{sec:policyhooks}, we suggest topic tags (e.g., pregnancy, asylum, labor) as privacy-preserving signals for stricter safeguards, rather than inferring protected attributes. This was not implemented in our experiments. As a design idea, it may also miss cases where risk is identity-conditioned but the query appears generic. \emph{Future work should explore how to operationalize such topic-aware safeguards and investigate richer, privacy-preserving signals that better capture identity-sensitive risks}.
    
    \item \textbf{Model Dependency:} Current findings hinge on specific LLMs (GPT-4.1 variants). As models evolve, the behavior of MetaRAG may shift. \emph{Future efforts should validate MetaRAG across open-source and emerging models to reinforce its robustness}.
    \item \textbf{Efficiency and Cost:} The verification steps add computational overhead, possibly impacting deployment in latency sensitive environments. \emph{Investigating lighter-weight verification strategies and adaptive scheduling techniques could help mitigate this trade-off}.
    \item \textbf{Context Modality:} Our current formulation assumes that the retrieved context $C$ is textual, enabling direct comparison through language-based verification. However, RAG pipelines increasingly operate over multimodal contexts such as tables, structured knowledge bases, or images. \emph{Future work should extend MetaRAG to handle non-textual evidence, requiring modality-specific verification strategies (e.g., table grounding, multimodal alignment)}.
\end{itemize}

Together, these limitations highlight both immediate boundaries and promising future directions for enhancing MetaRAG’s reliability, fairness, and efficiency.

\section{Conclusion}

Hallucinations in RAG-based conversational agents remain a significant barrier to trustworthy deployment in real-world applications.  
We introduced \textbf{MetaRAG}, a metamorphic testing framework for hallucination detection in retrieval-augmented generation (RAG) that operates without requiring ground truth references or access to model internals.  
Our experiments show that MetaRAG achieves strong detection performance on a challenging proprietary dataset. Beyond general reliability, MetaRAG’s factoid-level localization further supports identity-aware deployment by surfacing unsupported claims in sensitive domains (e.g., healthcare, migration, labor). Looking ahead, we see MetaRAG as a step toward safer and fairer conversational AI, where hallucinations are not only detected but also connected to safeguards that protect users in identity-sensitive contexts. This connection to identity-aware AI ensures that hallucination detection does not treat all users as homogeneous but provides safeguards that reduce disproportionate risks for identity-specific groups.
\section*{Acknowledgment}
This work was carried out during Channdeth’s internship at Forvia. 
The authors thank the Forvia team in Bercy, Paris, for their guidance and support.
\section*{Declaration on Generative AI}

During the preparation of this work, the authors used ChatGPT and Grammarly to improve the clarity and grammar of certain sentences and to rephrase text for better readability. After using this tool, the authors reviewed, edited, and verified all content to ensure accuracy and originality, and they take full responsibility for the publication’s content.

%%
%% Define the bibliography file to be used
\bibliography{references}

\appendix
\section*{A. Prompt Templates}

\subsection*{A.1. Factoid Decomposition Prompt}
We provide the exact template used to extract atomic factoids from model responses.

\begin{lstlisting}
export const factExtractionPrompt = (inputText: string) => `
You are a fact extraction assistant.
Your task is to extract all specific factual propositions from the given text.

Instructions:
1. Extract every distinct factual statement present in the input, even if the statement is incorrect, ambiguous, or nonsensical.
2. Each extracted proposition must be a complete, standalone sentence.
3. Each sentence must express only one atomic fact. (An atomic fact cannot be split into simpler factual statements.)
4. If a sentence contains multiple facts, split them into multiple atomic fact sentences.
5. Do not paraphrase, rewrite, summarize, interpret, infer, or judge any part of the input. Only extract and restate what is explicitly written.
6. Do not omit or correct any statements, regardless of their factual accuracy.
7. Output your answer as a JSON array of strings, with each element being one atomic factual sentence.

Example:
Input:
Marie Curie discovered polonium and radium, and Albert Einstein developed the theory of relativity in 1905.

Output:
[
  "Marie Curie discovered polonium.",
  "Marie Curie discovered radium.",
  "Albert Einstein developed the theory of relativity in 1905."
]

Now, extract atomic facts from this text:

Input:
${inputText}
`;
\end{lstlisting}

\subsection*{A.2. Mutation Generation Prompts}
We provide both synonym and antonym generation templates
\begin{lstlisting}
export const antonymPrompt = (
  count: Integer,
  question: string,
  factoid: string
) => `
You will be given a question and a factual answer (factoid).

Your task is to generate ${count} *negations* (contradictory statements) of the factoid, based on the context of the question.

Instructions:
- Each negation must directly contradict the factoid, focusing on what the question asks.
- Do not add new information not present in the factoid or question.
- Do not use double negations or wording that preserves the original meaning.
- Each negation must be a meaningful, grammatically correct sentence.
- Do not introduce unrelated facts.
- Ensure that each negation is relevant to the question's context.
- **Just return the sentences, one per line, without numbers or bullets, and nothing else.**

Example:
Question: Where was Einstein born?
Factoid: Einstein was born in Germany.
Good Antonym: Einstein was not born in Germany.
Bad Antonym: Einstein visited Germany. (not a contradiction)
Bad Antonym: Einstein was born in Austria. (adds new information)
Bad Antonym: Einstein was not not born in Germany. (double negation)
Bad Antonym: Was not born in Germany. (missing subject)

Question: ${question}

Factoid: ${factoid}
`;
\end{lstlisting}

\begin{lstlisting}
export const synonymPrompt = (
  count: Integer,
  question: string,
  factoid: string
) => `
You will be given a question and a factual answer (factoid).

Your task is to generate ${count} *synonyms* (paraphrased statements with the same meaning) of the factoid, based on the context of the question.

Instructions:
- Each output must be a single atomic factual claim (cannot be split into smaller facts).
- Use only information explicitly present in the question or factoid. Do not invent, infer, or add external knowledge.
- A correct synonym is a statement that means exactly the same thing as the factoid, even if the wording is different.
- Do not output partial phrases, keywords, or combine/split facts.
- Each synonym must be a complete, grammatically correct sentence.
- Just return the sentences, one per line, without numbers, bullets, or any other output.

Example:
Question: Where was Einstein born?
Factoid: Einstein was born in Germany.
Good Synonym: Germany is the country where Einstein was born.
Bad Synonym: Einstein visited Germany. (not equivalent)
Bad Synonym: Einstein was born. (incomplete)

Question: ${question}

Factoid: ${factoid}
`;
\end{lstlisting}
\subsection*{A.3. Factoid Verification Prompt}
The verification step compares mutated factoids against retrieved context using entailment/contradiction/neutral checks.
\begin{lstlisting}
export const verifyPrompt= (
  statement: string,
  context: string
) => `
You will be given a statement and passages that represent the ground truth.

Determine if the statement is supported by the passage, either explicitly or through clear implication.

Answer with one of the following **only**:  
- YES: if the statement is clearly and completely supported by the passages.  
- NO: if the statement is contradicted or directly refuted by the passages.  
- NOT SURE: if the passage does not contain enough information to confirm or deny the statement.
 
Respond with YES, NO, or NOT SURE. Then, in one short sentence, explain the reason for your answer.

Examples:

Passages (Ground Truth): "Alice was born in Paris and moved to New York at the age of five."
Statement: "Alice spent her early childhood in France."
Answer: YES. The passage states Alice was born in Paris, which is in France.

Passages (Ground Truth): "Bob has never visited Japan but plans to travel there next summer."
Statement: "Bob visited Japan last year."
Answer: NO. The passage says Bob has never visited Japan

Passages (Ground Truth): "Carol enjoys outdoor activities like hiking and cycling."
Statement: "Carol loves swimming."
Answer: NOT SURE. There is no information in the passages about Carol and swimming.

Now, perform the task:

Passages (Ground Truth): ${context}
Statement: ${statement}
Answer:`;
\end{lstlisting}
All verification LLMs were run with temperature $T=0.0$ to ensure determinism.  
\section*{B. Dataset and Annotation}

\subsection*{B.1. Dataset}

Our evaluation dataset consists of 23 internal enterprise documents, unseen during LLM training. Each document was segmented into chunks of approximately a few hundred tokens, and we retrieved up to $k=3$ chunks per query. Retrieval used cosine similarity over OpenAI \texttt{text-embedding-3-large}.

\begin{figure}[h]
    \centering
    \includegraphics[width=0.99\linewidth]{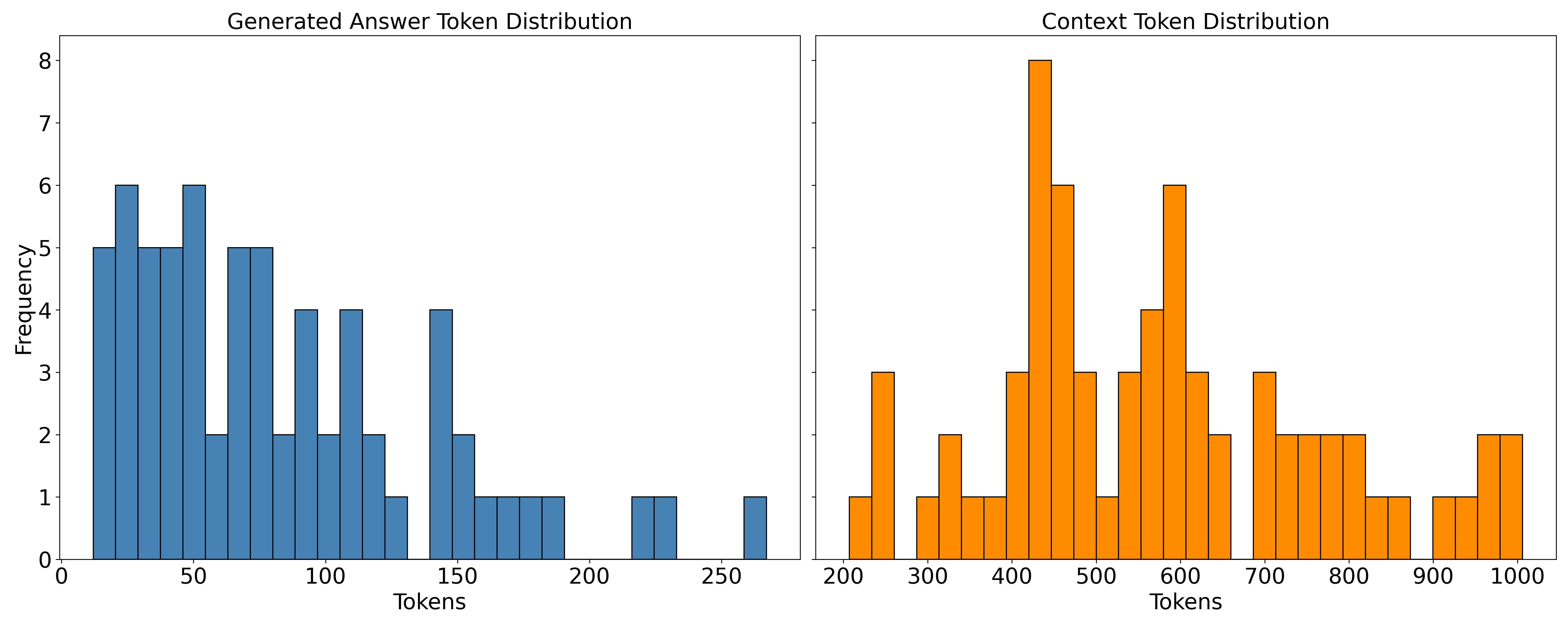}
    \caption{Token length distributions of generated answers (left) and retrieved context passages (right).}
    \label{fig:token_lengths}
\end{figure}

Figure~\ref{fig:token_lengths} further illustrates the token length distributions of generated answers and retrieved contexts. Generated answers are typically shorter (median $\approx$ 83 tokens), while retrieved contexts are longer (median $\approx$ 572 tokens), reflecting the compression and grounding challenges faced by the RAG system.

\subsection*{B.2. Human Annotation Protocol}
The annotation dataset was constructed in three steps. 
First, we collected responses produced by the RAG system on enterprise documents. 
Second, we used an LLM-based verifier to provide an initial label (\textbf{faithful} or \textbf{hallucinated}) for each response based on its retrieved context.
Finally, human annotators reviewed the RAG responses together with their retrieved evidence and assigned gold labels. Annotators were instructed to:  
\begin{itemize}
  \item Mark each response as \textbf{faithful} or \textbf{hallucinated}.  
  \item Consider a response hallucinated if any atomic factoid was not supported by retrieved evidence.  
  \item Resolve ambiguous cases by majority vote.  
\end{itemize}

\begin{figure}[h]
    \centering
    \includegraphics[width=0.7\linewidth]{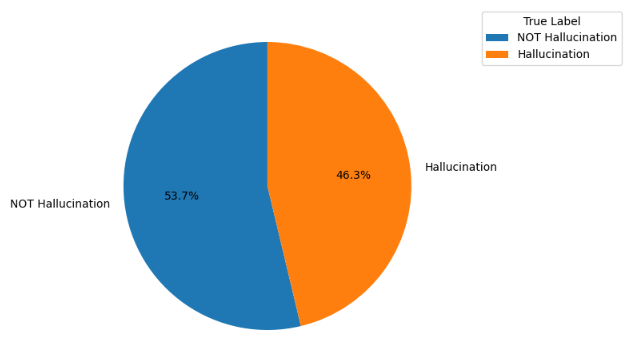}
    \caption{Proportion of hallucination vs. non-hallucination labels in the annotated dataset.}
    \label{fig:label_distribution}
\end{figure}

To ensure class balance across conditions, a subset of responses was lightly edited (e.g., by introducing or removing unsupported factual details) so that hallucinated and non-hallucinated examples were more evenly represented. These edits were applied before annotation, and annotators labeled both original and modified responses using the same guidelines.
Figure~\ref{fig:label_distribution} illustrates the final label distribution in our dataset, confirming that hallucination and not hallucination cases are reasonably balanced.

\section*{C. Extended Results}
\subsection*{C.1. Result}
Table~\ref{tab:ablation_full} reports the full ablation results across prompt settings, mutation counts, and verifier models.

\begin{table*}[ht]
  \centering
    \caption{Complete MetaRAG ablation results across 26 configurations. 
    Configuration format: Decomposition Model / Mutation Generation Model / Verifier Model / Number of Mutations ($N$) / Temperature. 
    \textit{Total (avg)} denotes the average execution time, and \textit{Cost (avg)} denotes the average token usage per run.}
    
  \label{tab:ablation_full}
  \scriptsize
  \setlength{\tabcolsep}{3pt}
  \begin{tabular}{c l c c c c c c}
    \toprule
    ID & Configuration & Precision & Recall & F1 & Accuracy & Time (avg) & Tokens (avg) \\
    \midrule
    0  & 41 / mini / mini/2/0.0 & 0.784 & 0.935 & 0.853 & 0.851 & 2.490 & 35.5k \\
    1  & mini / 41 / 41/2/0.7  & 0.776 & 1.000 & 0.874 & 0.866 & 3.542 & 29.0k \\
    2  & mini / 41 / multi /2/0.7& 0.828 & 0.935 & 0.878 & 0.880 & 3.205 & 28.8k \\
    3  & mini / 41 / mini/2/0.7& 0.806 & 0.935 & 0.866 & 0.866 & 3.011 & 28.7k \\
    4  & mini / 41 / 41/2/0.0 & 0.757 & 1.000 & 0.861 & 0.851 & 2.705 & 28.9k \\
    5  & mini / 41 / multi/2/0.0 & 0.885 & 1.000 & 0.939 & 0.940 & 2.287 & 29.1k \\
    6  & mini / 41 / mini/2/0.0& 0.784 & 0.935 & 0.853 & 0.851 & 2.581 & 29.0k \\
    7  & mini / mini / 41/2/0.7 & 0.764 & 0.935 & 0.841 & 0.836 & 2.064 & 29.1k \\
    8  & mini / mini /multi/2/0.7& 0.828 & 0.935 & 0.878 & 0.880 & 2.033 & 28.9k \\
    9  & mini / mini / mini/2/0.7& 0.837 & 1.000 & 0.911 & 0.910 & 1.814 & 29.1k \\
    10 & 41 / 41 / mini/2/0.0& 0.790 & 0.968 & 0.870 & 0.866 & 3.471 & 36.0k \\
    11 & 41 / mini / 41/2/0.0   & 0.784 & 0.935 & 0.853 & 0.851 & 3.338 & 36.7k \\
    12 & 41 / mini / multi /2/0.0   & 0.764 & 0.935 & 0.841 & 0.836 & 3.204 & 37.1k \\
    13 & mini / mini / 41/2/0.0 & 0.757 & 0.903 & 0.824 & 0.821 & 2.200 & 28.7k \\
    14 & mini / mini / multi/2/0.0  & 0.828 & 0.935 & 0.878 & 0.880 & 2.834 & 28.7k \\
    15 & mini / mini / mini/2/0.0& 0.812 & 0.968 & 0.883 & 0.881 & 2.267 & 29.1k \\
    16 & mini / 41 / multi/5 /0.0    & 0.882 & 0.968 & 0.923 & 0.925 & 7.671 & 76.5k \\
    17 & mini / 41 / mini/5/0.0& 0.812 & 0.968 & 0.883 & 0.881 & 5.693 & 74.0k \\
    18 & mini / mini / 41/5/0.7 & 0.909 & 0.968 & 0.937 & 0.940 & 5.185 & 74.6k \\
    19 & mini / mini / 41/5 /0.0 & 0.935 & 0.935 & 0.935 & 0.940 & 5.215 & 77.5k \\
    20 & mini / mini / mini/5/0.7 & 0.866 & 0.838 & 0.852 & 0.865 & 3.710 & 77.4k \\
    21 & mini / mini / multi/5/0.7  & 0.896 & 0.838 & 0.866 & 0.880 & 5.903 & 80.2k \\
    22 & mini / mini / multi/5/0.0 & 0.964 & 0.870 & 0.915 & 0.925 & 5.812 & 76.2k \\
    23 & mini / mini / mini/5/0.0 & 0.874 & 0.903 & 0.888 & 0.895 & 3.515 & 75.1k \\
    24 & 41 / mini / multi /5/0.0 & 0.900 & 0.870 & 0.885 & 0.895 & 6.345 & 89.8k \\
    25 & 41 / mini / mini/5/0.0& 0.833 & 0.806 & 0.819 & 0.835 & 4.631 & 88.2k \\
    \bottomrule
  \end{tabular}
\end{table*}
\subsection*{C.2. Consistency Study }
To assess the robustness of MetaRAG to random initialization, we report mean and standard deviation of the main evaluation metrics over 5 different random seeds for the top configurations. We also include the coefficient of variation (CV) for Precision and Recall, which provides a normalized measure of variability relative to the mean.
\begin{table}[ht]
    \centering
    \caption{Run-to-run consistency for top precision configurations (mean $\pm$ standard deviation over 5 seeds) and coefficient of variation (CV) for Precision.}
    \label{tab:cv_table_precision}
    \scriptsize
    \begin{tabular}{lcccc}
        \toprule
        \textbf{ID} & \textbf{F1} & \textbf{Precision} & \textbf{Recall} & \textbf{CV (Prec.)} \\
        \midrule
        19 & 0.9347 $\pm$ 0.0305 & 0.9410 $\pm$ 0.0357 & 0.9286 $\pm$ 0.0272 & 3.79\% \\
        18 & 0.9356 $\pm$ 0.0089 & 0.9322 $\pm$ 0.0439 & 0.9413 $\pm$ 0.0278 & 4.71\% \\
        22 & 0.8928 $\pm$ 0.0351 & 0.9246 $\pm$ 0.0359 & 0.8641 $\pm$ 0.0478 & 3.88\% \\
        24 & 0.8911 $\pm$ 0.0272 & 0.9064 $\pm$ 0.0144 & 0.8772 $\pm$ 0.0475 & 1.59\% \\
        \bottomrule
    \end{tabular}
\end{table}

\begin{table}[ht]
    \centering
    \caption{Run-to-run consistency for top recall configurations (mean $\pm$ standard deviation over 5 seeds) and coefficient of variation (CV) for Recall.}
    \label{tab:cv_recall_table}
    \scriptsize
    \begin{tabular}{lcccc}
        \toprule
        \textbf{ID} & \textbf{F1} & \textbf{Precision} & \textbf{Recall} & \textbf{CV (Recall)} \\
        \midrule
        1 & 0.8910 $\pm$ 0.0298 & 0.8045 $\pm$ 0.0496 & 1.0000 $\pm$ 0.0000 & 0.00\%  \\
        4 & 0.8756 $\pm$ 0.0217 & 0.7829 $\pm$ 0.0283 & 0.9935 $\pm$ 0.0145 & 1.46\% \\
        5 & 0.9108 $\pm$ 0.0346 & 0.8463 $\pm$ 0.0503 & 0.9869 $\pm$ 0.0179 & 1.82\%  \\
        9 & 0.8623 $\pm$ 0.0434 & 0.7910 $\pm$ 0.0414 & 0.9482 $\pm$ 0.0491 & 5.18\% \\
        \bottomrule
    \end{tabular}
\end{table}

These results (Table~\ref{tab:cv_table} and Table~\ref{tab:cv_recall_table}) demonstrate that MetaRAG maintains stable performance across random seeds, particularly for high-precision configurations (e.g., IDs 24) and high-recall configurations (e.g., IDs 1, 4 and 5). This stability supports the reliability of the Pareto front analysis presented in the following section.

\subsection*{C.3. Pareto Front Analysis}

\begin{figure*}[t]
    \centering
    \begin{subfigure}{0.9\linewidth}
        \centering
        \includegraphics[width=\linewidth]{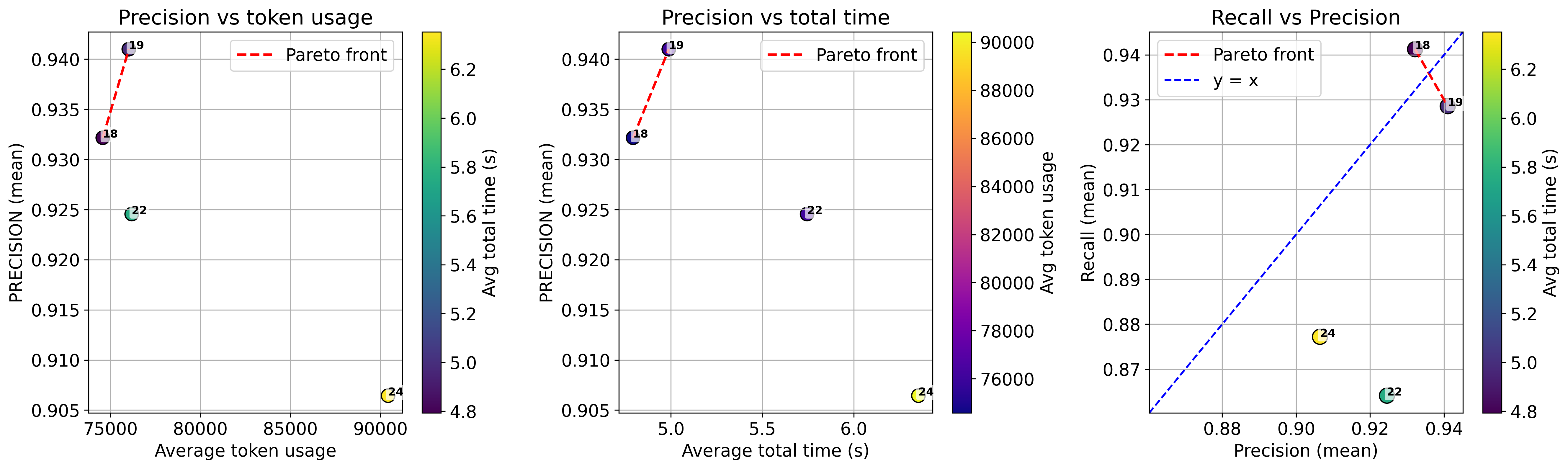}
        \caption{Precision Pareto front (vs. cost/latency)}
        \label{fig:pareto-precision}
    \end{subfigure}
    
    \vspace{0.5em}
    
    \begin{subfigure}{0.9\linewidth}
        \centering
        \includegraphics[width=\linewidth]{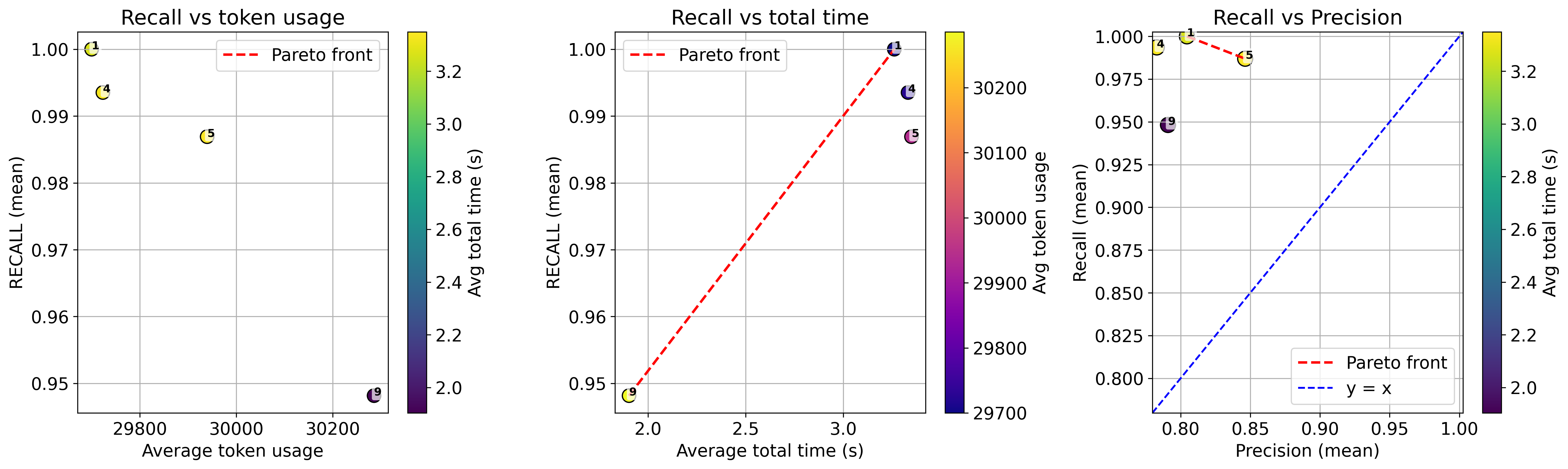}
        \caption{Recall Pareto front (vs. cost/latency)}
        \label{fig:pareto-recall}
    \end{subfigure}
    \vspace{1em}
    \caption{Pareto front analysis for retrieval budget. Each point corresponds 
    to a MetaRAG configuration; non-dominated (Pareto-optimal) points are 
    highlighted. Subfigure~(a) shows Precision trade-offs, and Subfigure~(b) 
    shows Recall trade-offs.}
    \label{fig:pareto-front}
\end{figure*}

We further analyze robustness and metric-specific trade-offs in this 
Supplementary Material. A configuration is Pareto-optimal if no other 
configuration achieves strictly higher performance while being no worse 
in the cost metrics. Figures~\ref{fig:pareto-precision} and~\ref{fig:pareto-recall} 
present the corresponding Pareto fronts for Precision and Recall. 
These analyses confirm that the same top-ranked configurations 
(IDs~5, 18, 19, and~16) consistently offer strong performance–efficiency 
trade-offs across multiple evaluation criteria.

In our setting, the positive class corresponds to hallucinations, while the 
negative class corresponds to faithful responses (no hallucinations). Hence, \textbf{high Precision }
means that flagged hallucinations are rarely false positives, which is 
critical in \textit{safety-critical} and \textit{trustworthy applications}. Conversely, 
\textbf{high Recall} ensures that most hallucinations are detected, though at 
the cost of occasionally misclassifying faithful responses. Such recall-oriented 
configurations may be advantageous in exploratory or diagnostic scenarios. In practice, high-precision operating points (e.g., IDs~18 and~19) reduce false alarms in safety-critical pipelines, while high-recall points (e.g., IDs~1 and~4) maximize coverage in exploratory settings. This mirrors standard alert-system trade-offs and clarifies how \textbf{MetaRAG} can be tuned for different deployment objectives. Selections based on F1 score represent a balanced compromise suitable for general-purpose use cases.

\subsection*{D. Implementation Notes}
All \textbf{MetaRAG} experiments were implemented in TypeScript with asynchronous API calls to the LLMs, allowing multiple requests to be processed concurrently. This parallelization reduced the average end-to-end execution time per run, without affecting accuracy metrics. The reported runtime and cost results in Table~\ref{tab:ablation_full} are therefore representative of a practical and scalable setup.

\end{document}